\documentclass[letterpaper, 10 pt, conference]{ieeeconf}

\IEEEoverridecommandlockouts                              

\usepackage{graphics} 
\usepackage{amsmath,amssymb,amsfonts} 
\usepackage{bm}
\usepackage{subcaption}	 
\usepackage{graphicx}
\usepackage{multirow}
\usepackage{array}
\usepackage{caption}

\usepackage{url}
\usepackage{hyperref}
\hypersetup{colorlinks,urlcolor=red}
\usepackage{xcolor}

\usepackage{colortbl}

\usepackage{makecell}
\usepackage{arydshln}

\usepackage{graphicx}
\usepackage{caption}
\usepackage{subcaption}
\usepackage{float}

\usepackage{booktabs}
\usepackage{pifont}

\usepackage[backend=bibtex,natbib=true,style=numeric-comp,sorting=none,giveninits=true,maxbibnames=99,url=false,doi=false]{biblatex}
\title{\LARGE \bf
Gaussian-LIC: Real-Time Photo-Realistic SLAM with \\ Gaussian Splatting and LiDAR-Inertial-Camera Fusion 
}

\author{Xiaolei Lang$^{1,\dagger}$, Laijian Li$^{1,\dagger}$, Chenming Wu$^{2}$, Chen Zhao$^{2}$, Lina Liu$^{1}$, 
Yong Liu{$^{1}$}, Jiajun Lv{$^{1,\ast}$}, Xingxing Zuo{$^{3,\ast}$}
\thanks{$^{1}$Institute of Cyber-Systems and Control, Zhejiang University, China.}%
\thanks{$^{2}$Baidu VIS, China.}%
\thanks{$^{3}$California Institute of Technology, USA.}%
\thanks{$^\dag$Contributed equally. $^\ast$Corresponding authors.}
}

\begin{document}

\maketitle
\thispagestyle{empty}
\pagestyle{empty}

\begin{abstract}
In this paper, we present a real-time photo-realistic SLAM method based on marrying Gaussian Splatting with LiDAR-Inertial-Camera SLAM. Most existing radiance-field-based SLAM systems mainly focus on bounded indoor environments, equipped with RGB-D or RGB sensors. However, they are prone to decline when expanding to unbounded scenes or encountering adverse conditions, such as violent motions and changing illumination. In contrast, oriented to general scenarios, our approach additionally tightly fuses LiDAR, IMU, and camera for robust pose estimation and photo-realistic online mapping. To compensate for regions unobserved by the LiDAR, we propose to integrate both the triangulated visual points from images and LiDAR points for initializing 3D Gaussians. In addition, the modeling of the sky and varying camera exposure have been realized for high-quality rendering. Notably, we implement our system purely with C++ and CUDA, and meticulously design a series of strategies to accelerate the online optimization of the Gaussian-based scene representation. Extensive experiments demonstrate that our method outperforms its counterparts while maintaining real-time capability. 
Impressively, regarding photo-realistic mapping, our method with our estimated poses even surpasses all the compared approaches that utilize privileged ground-truth poses for mapping.
Our code will be released on project page \url{https://xingxingzuo.github.io/gaussian_lic}.
\end{abstract}

\section{INTRODUCTION}

Simultaneous localization and mapping (SLAM) is a pivotal technique in mixed reality and robotics. In recent years, emerging radiance field technologies such as Neural Radiance Fields (NeRF)~\cite{mildenhall2021nerf}  and 3D Gaussian Splatting (3DGS)~\cite{kerbl20233d} have carved out a new direction in the realm of SLAM, namely radiance-field-based SLAM~\cite{tosi2024nerfs}, aiming to simultaneously localize accurately and construct photo-realistic 3D maps. This holds significant importance for enhancing the capabilities of robots to model, understand, and interact with natural environments in real time~\cite{zheng2024gaussiangrasper, jin2024gs, chen2024splat, zuo2024fmgs}. 

Compared to NeRF which involves computationally intensive volumetric rendering based on ray tracing, 3DGS offers fast rendering speed with superior visual quality, showcasing a greater potential in real-time SLAM applications. Yet there still remain several issues for 3DGS-based SLAM. First, similar to existing NeRF-based SLAM~\cite{sucar2021imap, zhu2022nice, yang2022vox, wang2023co, johari2023eslam, sandstrom2023point, xin2024hero, rosinol2023nerf, naumann2023nerf}, most 3DGS-based SLAM~\cite{yan2024gs, keetha2024splatam, yugay2023gaussian, ha2024rgbd, hu2024cg, peng2024rtg, sarikamis2024ig, zhu2024mgs, huang2024photo, matsuki2024gaussian, li2024sgs, zhu2024semgauss, ji2024neds} focus mainly on indoor scenarios using RGB-D or RGB sensors. When it comes to unbounded outdoor scenes or encountering challenging conditions such as violent ego-motion, varying illumination, and lack of visual textures, the performance of existing 3DGS-based SLAM might dramatically diminish. Second, for the outdoor scenarios, the modeling of the sky and varying camera exposure becomes important. Ignoring them can incur significant artifacts to 3DGS in unbounded scenes. 
Third, real-time performance is still a bottleneck. Constraining the number of or the dimension of attributes of 3D Gaussian representation~\cite{keetha2024splatam, peng2024rtg} could be effective indoors but might lead to deteriorated performance in unbounded outdoor scenarios. 

To address the problems above, we propose Gaussian-LIC, a photo-realistic SLAM system that tightly fuses LiDAR-Inertial-Camera data for robust pose tracking and photo-realistic reconstruction with 3D Gaussian map representation \textbf{all in real time}. The 3D Gaussians can be initialized using both the LiDAR points and visual SFM (struct-from-motion) points (obtained by online triangulation). Both the sky and changing camera exposure are taken into consideration in our system for higher-quality reconstruction. Importantly, we carefully design acceleration strategies for Gaussian optimization in CUDA implementation. In summary, our main contributions are as follows:
\begin{itemize}
	\item We present the \textbf{first} real-time photo-realistic SLAM with Gaussian Splatting and LiDAR-Inertial-Camera fusion in unbounded outdoor scenarios. Our SLAM framework delivers robust pose estimation while incrementally constructing a high-fidelity Gaussian map, all without the need for extensive post-processing after the sensory data is received.
	\item For high-quality rendering in broad 3D space, we seamlessly incorporate LiDAR points and 3D visual landmarks obtained from triangulation for initialization of 3D Gaussians. Also, we model the sky and varying camera exposure to better tackle the complex unbounded scenes outdoors.
    \item We implement our system purely in C++ and CUDA and meticulously design a series of CUDA-related acceleration strategies to boost the training of Gaussians, which will be open-sourced to benefit the community.
    
	\item We conduct comprehensive experiments on real-world indoor and outdoor datasets using various types of LiDAR, demonstrating that our method achieves state-of-the-art performance in real-time photorealistic mapping, even under adverse conditions. Remarkably, our approach, using our estimated poses, even outperforms all compared methods that rely on privileged ground-truth poses for mapping.
\end{itemize}

\section{RELATED WORK}

\subsection{Photo-Realistic RGB-D or RGB SLAM}
Given sequential RGB-D or RGB inputs, abundant works attain brilliant results of localization and photo-realistic mapping in indoor environments. iMAP~\cite{sucar2021imap}, the first NeRF-based SLAM that employs the implicit neural representation to achieve watertight online reconstruction, has pioneered a new era in SLAM. As a follow-up, NICE-SLAM~\cite{zhu2022nice} combines MLPs with hierarchical feature grids,  markedly improving the quality of scene representation and realizing outstanding performance in larger indoor rooms. Further, Vox-Fusion~\cite{yang2022vox} utilizes octree to dynamically expand the volumetric neural implicit map, eliminating the need for pre-allocated grids. Adopting hash-grids, tri-planes, and neural point clouds as implicit neural representations respectively, Co-SLAM~\cite{wang2023co}, ESLAM~\cite{johari2023eslam}, and Point-SLAM~\cite{sandstrom2023point} get enhancement in both localization and reconstruction. Meanwhile, HERO-SLAM~\cite{xin2024hero} improves the robustness in the optimization of the neural map. NeRF employs time-consuming ray-based volume rendering, while 3DGS utilizes fast point-based rasterization, accelerating image view synthesis and producing promising rendering quality. Notably, GS-SLAM~\cite{yan2024gs}, SplaTAM~\cite{keetha2024splatam}, and Gaussian-SLAM~\cite{yugay2023gaussian} detailedly elucidate the significant advantages of 3DGS over existing map representations in SLAM tasks for online photo-realistic mapping. SGS-SLAM~\cite{li2024sgs}, SemGauss-SLAM~\cite{zhu2024semgauss}, and NEDS-SLAM~\cite{ji2024neds} integrate semantic feature embedding into Gaussians and demonstrate superb capabilities in online dense semantic mapping. Fusing Generalized Iterative Closest Point (G-ICP) and 3DGS, GS-ICP-SLAM~\cite{ha2024rgbd} shares covariances of Gaussians between tracking and mapping to minimize redundant computations. CG-SLAM~\cite{hu2024cg} additionally incorporates a depth uncertainty model to select valuable Gaussians and thereby improves tracking performance. By forcing each Gaussian to be either opaque
or nearly transparent, RTG-SLAM~\cite{peng2024rtg} achieves real-time performance indoors with compact scene representation.

Several works have tried to operate solely on monocular images. Among them, NeRF-SLAM~\cite{rosinol2023nerf} and IG-SLAM~\cite{sarikamis2024ig} utilize the dense depth maps estimated from the tracking front-end DROID-SLAM~\cite{teed2021droid} as additional information to supervise the training of Instant-NGP~\cite{muller2022instant} and 3DGS. Correspondingly, NeRF-VO~\cite{naumann2023nerf} and MGS-SLAM~\cite{zhu2024mgs} instead employ the sparse visual odometry DPVO~\cite{teed2024deep} as a faster front-end with network-predicted dense depth for supervision. Concurrently, Photo-SLAM~\cite{huang2024photo} utilizes the classical visual odometry ORB-SLAM3~\cite{campos2021orb} for accurate pose estimation and reconstructs a hybrid Gaussian map with ORB features. MonoGS~\cite{matsuki2024gaussian} introduces geometric regularization to address ambiguities in incremental reconstruction.

\subsection{Photo-Realistic Multimodal SLAM}
RGB-D and RGB approaches can realize fancy results in well-lit indoor scenes, but they would find it difficult to scale up to complex unbounded outdoor environments or cope with violent motions, lighting changes, and texture missing. Also, RGB-D sensors are often unavailable outdoors. To this end, fusing multimodal sensors such as LiDAR and IMU will be necessary and effective, which has been extensively proven in classical SLAM methods~\cite{zuo2019lic, zuo2020lic, shan2021lvi, lin2022r, zheng2022fast, lv2023continuous, lang2023coco} characterized by robustness and accuracy. Nevertheless, their primary focus lies in geometric mapping, resulting in geometrically precise yet visually simple representations of the scene. URF~\cite{rematas2022urban}, DrivingGaussian~\cite{zhou2024drivinggaussian}, StreetGaussians~\cite{yan2024street}, PVG~\cite{chen2023periodic}, LIV-GaussMap~\cite{hong2024liv}, and LetsGo~\cite{cui2024letsgo} perform enhanced photorealistic reconstruction based on LiDAR-camera
fusion, but all of them are offline methods taking a batch input. 

Oriented to SLAM tasks, MM-Gaussian~\cite{wu2024mm} leverages accurate geometric structure information from a solid-state LiDAR to substitute the depth of the RGB-D, thus scaling up to the outdoor scenes. Specifically, it first performs a pure LiDAR odometry~\cite{vizzo2023kiss} for pose estimations and further refines the pose via color and depth rendering loss. Similar to SplaTAM~\cite{keetha2024splatam}, it optimizes isotropic Gaussians without view-dependent effect initialized from LiDAR points. MM3DGS-SLAM~\cite{sun2024mm3dgs} fuses RGB-D camera and IMU measurements in a loosely-coupled manner, to enable more accurate and scale-aware pose estimation. Both of the aforementioned multimodal SLAM approaches are capable of functioning in outdoor environments. However, their mapping performance is suboptimal due to the lack of sky modeling and the challenges posed by varying camera exposure in outdoor settings. Additionally, they are not specially optimized for real-time applications.

\section{METHODOLOGY}
\label{sec:method}

\begin{figure*}[t]
    	\centering
    	\includegraphics[width=\textwidth]{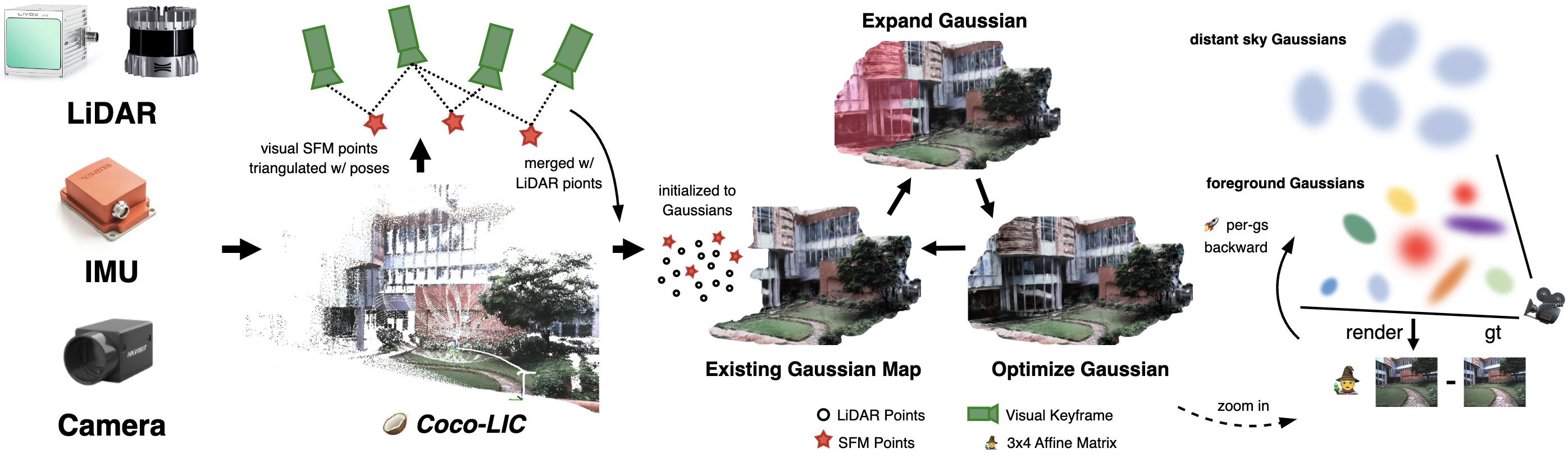}

    	\caption{The pipeline of our novel photo-realistic LiDAR-Inertial-Camera SLAM using 3D Gaussian Splatting. 
     }
    \label{fig:pipeline}
    \vspace{-1em}
    \end{figure*}

Our proposed Gaussian-LIC consists of two main modules: LiDAR-Inertial-Camera odometry with mapping data preparation (Sec.~\ref{sec:coco-lic}) and 3DGS-based photo-realistic online mapping (Sec.~\ref{sec:garlic-mapping}) powered by the carefully designed acceleration strategies (Sec.~\ref{sec:acceleration}) for real-time performance. Fig.~\ref{fig:pipeline} shows an overview of our system.

\subsection{Preliminary: 3D Gaussian Splatting}
\label{sec:3dgs}
To reconstruct a richly detailed dense map for fast and high-quality rendering, we represent the scene as a collection of anisotropic 3D Gaussians~\cite{kerbl20233d}, each of which contains position $\boldsymbol{\mu}\in\mathbb{R}^{3}$ in the world coordinate, scale $\mathbf{S}\in\mathbb{R}^{3}$, rotation $\mathbf{R}\in\mathbb{R}^{3 \times 3}$, opacity $o\in\mathbb{R}$, and three-degree spherical harmonics $\mathbf{SH}\in\mathbb{R}^{48}$ capturing the view-dependent appearance of the scene. Representing the Gaussian’s ellipsoidal shape, the covariance of each Gaussian is parameterized as:
\begin{align}
    \boldsymbol{\Sigma} = \mathbf{R}\mathbf{S}\mathbf{S}^T\mathbf{R}^T.
\end{align}

Given the camera pose ${}^C_W\mathbf{T}=\{{}^C_W\mathbf{R}, {}^C\mathbf{p}_W\}$ which transforms the point ${}^W\mathbf{p}$ in the world frame $\{W\}$ into the camera frame $\{C\}$, we can splat a 3D Gaussian $\mathcal{N}\left(\boldsymbol{\mu}, \boldsymbol{\Sigma}\right)$ onto the image screen and get a 2D Gaussian $\mathcal{N}\left(\boldsymbol{\mu}^{\prime}, \boldsymbol{\Sigma}^{\prime}\right)$:
\begin{gather}
    \boldsymbol{\mu}^{\prime} = \pi_c\left( \frac{\hat{\boldsymbol{\mu}}}{\mathbf{e}_{3}^{\top} \hat{\boldsymbol{\mu}}} \right), \ \hat{\boldsymbol{\mu}} = {}^C_W\mathbf{R} \; \boldsymbol{\mu} + {}^C\mathbf{p}_W\,,
    \\
    \boldsymbol{\Sigma}^{\prime} = [\mathbf{J} \; {}^C_W\mathbf{R} \; \boldsymbol{\Sigma}  \; {{}^C_W\mathbf{R}}^T\mathbf{J}^T]_{2\times 2}\,,
\end{gather}
where $\mathbf{e}_{i}$ is a $3\times1$ vector with its $i$-th element to be 1 and the other elements to be 0. Then ${\mathbf{e}_{3}^{\top} \hat{\boldsymbol{\mu}}}$ is the depth $d$ of the Gaussian in the camera frame. $\pi_c(\cdot)$ denotes the projection operation which transforms a point on the normalized image plane to a pixel. $\mathbf{J}$ is the Jacobian of the affine approximation of the projective transformation and $[\cdot]_{2\times 2}$ skips the third row and column to acquire a $2\times 2$ matrix. The projected 2D Gaussian affects the pixel $\boldsymbol{\rho} = \begin{bmatrix} u & v \end{bmatrix}^\top$ by the weight:
\begin{align}
    \alpha = o\exp\left(-\frac{1}{2}(\boldsymbol{\mu}^{\prime}-\boldsymbol{\rho})^T(\boldsymbol{\Sigma}^{\prime})^{-1}(\boldsymbol{\mu}^{\prime}-\boldsymbol{\rho})\right)\,.
\end{align}

Sorting the Gaussians in depth order, the color of a pixel $\boldsymbol{\rho}$ can be efficiently rendered via front-to-back $\alpha$-blending:
\begin{align}
\label{eq:color-loss}
    \mathbf{C}(\boldsymbol{\rho})=\sum_{i=1}^n\mathbf{c}_i\alpha_i\prod_{j=1}^{i-1}(1-\alpha_j),
\end{align}
where $\mathbf{c}$ denotes the RGB color of the Gaussian computed from $\mathbf{SH}$. Similarly, we also render the depth map and opacity map by: 
\begin{gather}
    \mathbf{D}(\boldsymbol{\rho})=\sum_{i=1}^nd_i\alpha_i\prod_{j=1}^{i-1}(1-\alpha_j), \\
\mathbf{O}(\boldsymbol{\rho})=\sum_{i=1}^n\alpha_i\prod_{j=1}^{i-1}(1-\alpha_j).
\label{eq:silhouette}
\end{gather}

\subsection{Multimodal Odometry and Mapping Data Preparation}
\label{sec:coco-lic}
Given data from multimodal sensors from an RGB camera, a 3D LiDAR, and an IMU, we adopt our previous work Coco-LIC~\cite{lang2023coco}, a continuous-time odometry to tightly fuse the sensor data, delivering robust, real-time, and accurate pose estimation even in challenging scenarios. It incrementally optimizes the trajectory for every 0.1 seconds via factor graph optimization, incorporating point-to-map LiDAR factors, frame-to-map visual factors, and inertial factors.

For photo-realistic reconstruction, 3D Gaussians in the map need to be bootstrapped by a set of 3D points. The better initialization leads to faster convergence and higher-quality rendering. Exactly, the LiDAR points provide handy and more precise geometric priors for Gaussian initialization~\cite{zhou2024drivinggaussian}. Our system supports both solid-state and mechanical spinning LiDARs.
Specifically, after finishing the sliding-window optimization of the trajectory segment in the latest 0.1 seconds, we first downsample the LiDAR points in this time interval by retaining one out of every $N_l$ points randomly. Then we colorize the remained LiDAR points by projection onto the latest image and remove those outside the corresponding camera FoV (field of view). Afterward, these colorized LiDAR points in the world frame are sent to the mapping thread, together with the posed latest image, regarded as a frame.

Though offering accurate geometric priors, the region scanned by the LiDAR is limited compared to the camera, especially for the repetitive mechanical spinning LiDAR. Thus, we propose maintaining a visual sliding window similar to VINS-Mono~\cite{qin2018vins} to compute triangulated SFM points for compensation. The window consists of $N_k$ keyframes carefully selected from the latest images based on co-visibility, and every keyframe contains detected Shi-Tomasi corners~\cite{shi1994good} tracked by the KLT sparse optical flow~\cite{lucas1981iterative}. We triangulate a corner with the estimated keyframe poses and camera intrinsics once it is consecutively tracked in $N_t$ keyframes. The newly triangulated SFM points will be combined with LiDAR points for supplementation.

\subsection{Photo-Realistic Mapping using Gaussian Splatting}
\label{sec:garlic-mapping}
The mapping module receives sequential frames from the tracking module to perform reconstruction. Each frame consists of an estimated camera pose, an undistorted image, and colored LiDAR points combined with visual SFM points.

\subsubsection{Initialization of 3D Gaussians}
\label{sec:initialization}
The photo-realistic Gaussian map is bootstrapped with all 3D points of the first received frame. To be specific, for each point, we initialize a new Gaussian centered at its position with the zeroth degree of $\mathbf{SH}$ filled with its RGB color, opacity set to be $o_a$, and rotation set to be the identity matrix. To mitigate the aliasing artifacts~\cite{yan2024multi}, we assign smaller scales for the closer Gaussians, while larger scales for those far away from the image plane, namely $\frac{d}{f} \mathbf{e}$, where $\mathbf{e}$ is a $3\times1$ vector filled with 1, while $d$ and $f$ denote the depth of the point in the current camera frame and the camera focal length, respectively. 

In unbounded outdoor scenes, improper modeling of the sky would enforce Gaussians on the object surfaces to incorrectly fit the appearance of the sky, undermining the rendering quality. Thus we initialize a batch of sky Gaussians at random $N_s$ positions on the upper hemisphere specified by the world coordinate system as the center and a large radius $R$. The zeroth degree of $\mathbf{SH}$ of each sky Gaussian is filled with white color and the opacity is set to be $o_b$. Also, the initial scales of the sky Gaussians are determined by the closest distances between them. Consequently, these background sky Gaussians fall behind the foreground Gaussians. 

\subsubsection{Expansion of Gaussian Map}
\label{sec:expansion}
Every frame after the first usually captures
the geometry and appearance of the newly observed areas.
However, LiDAR points from different frames may contain
duplicate information. We take every fifth frame as a keyframe. When a keyframe is received, we first merge all points received in these five frames into a point cloud. To avoid redundancy, we then render $\mathbf{O}$ from the current keyframe image view according to Eq.~\eqref{eq:silhouette} and generate a mask $\mathbf{M}$ to select pixels that are not reliable from the current Gaussian map and tend to observe new areas:
\begin{align}
    \mathbf{M} = \mathbf{O} < \tau.
\end{align}

Only points in the merged point cloud that can be projected onto the selected pixels are utilized to initialize new Gaussians as  Sec.~\ref{sec:initialization} does, to expand the Gaussian map. 

\subsubsection{Optimization}
Once the current received frame is a keyframe, we randomly select $K$ keyframes from all keyframes to optimize the Gaussian map, avoiding the catastrophic
forgetting problems and maintaining the geometric consistency
of the global map. We randomly shuffle the selected $K$ keyframes and iterate through each of them to optimize the map by minimizing re-rendering loss: 
\begin{align}
    \mathcal{L}=(1-\lambda) \left\|\mathbf{E}[\mathbf{C}]-\bar{\mathbf{C}}\right\|_1+\lambda \mathcal{L}_{\mathrm{D}-\mathrm{SSIM}},
\end{align}
where $\bar{\mathbf{C}}$ and $\mathbf{C}$ are the ground-truth image and the image rendered by Eq.~\eqref{eq:color-loss}, respectively. $\mathcal{L}_{\mathrm{D}-\mathrm{SSIM}}$ is a D-SSIM term~\cite{kerbl20233d}. To handle the varying camera exposure, we additionally optimize a $3 \times 4$ affine matrix $\mathbf{E}$ which is applied to the rendered color~\cite{kerbl2024hierarchical}. The first three columns of $\mathbf{E}$ are used for scaling, and the last column is used for offsetting.

\subsection{Meticulous CUDA-related Designs for Acceleration}
\label{sec:acceleration}
Implemented fully in C++ and CUDA, our SLAM system consists of a tracking module and a mapping module, running in parallel and exchanging data based on ROS. The mapping module is based on the LibTorch framework. Inspired by~\cite{mallick2024taming, radl2024stopthepop}, we carefully integrate and design a series of strategies related to CUDA to boost the training of Gaussians, an important aspect overlooked in previous works.   
\subsubsection{Per-Gaussian Backpropagation}
During the backward pass of Gaussian Splatting~\cite{kerbl20233d}, gradients are propagated from pixels to Gaussians, constituting the most computationally expensive part of training. Specifically, each pixel corresponds to a varying number of Gaussians, and the overall time cost is dictated by the pixel with the highest number of Gaussians. Consequently, all other CUDA threads remain idle, waiting for this bottleneck pixel to complete its backpropagation. Additionally, atomic gradient addition conflicts arise when multiple pixels backpropagate to the same Gaussian. To mitigate this, we adopt a per-Gaussian backpropagation strategy~\cite{mallick2024taming}, assigning CUDA threads to all Gaussians within a tile. This allows to enhance parallelism while reducing collisions.

\subsubsection{Tile Culling and Additional Tricks}
The vanilla 3DGS~\cite{kerbl20233d} approximates projected 2D splatted Gaussians as circles to determine relevant tiles. However, this approach is suboptimal for slim 2D Gaussians, which frequently appear in incremental mapping systems such as SLAM, as they contribute minimally to most tiles selected by a circular approximation. In our implementation, we efficiently cull these weakly affected tiles through load balancing~\cite{radl2024stopthepop}, enhancing computational efficiency. For faster training of the Gaussian map representation, we further make use of the sparse Adam~\cite{mallick2024taming} to only update Gaussians in the current camera frustum instead of optimizing all the Gaussians.

\section{EXPERIMENTS}

\begin{table*}[htb]
\centering
\scriptsize
\caption{Quantitative rendering performance on public multimodal datasets. The best results are highlighted in bold. All the compared methods except ours utilize either the ground-truth poses or our estimated poses.} 
\begin{tabular}{llcccccccccccr}
\toprule
Method & Metric & \texttt{f0} & \texttt{f1} & \texttt{f2} & \texttt{r0} & \texttt{r1} & \texttt{r2} & \texttt{m0} & \texttt{m1} & \texttt{m2} & \texttt{Avg.} \\ 
\cmidrule(lr){1-12}
\multicolumn{12}{l}{\cellcolor[HTML]{EEEEEE}{\textit{Train View}}} \\ 

\multirow{3}{*}{\makecell[l]{NeRF-SLAM~\cite{rosinol2023nerf}}} 
& PSNR$\uparrow$  & 25.56 & 25.47 & 17.01 & 19.53 & 21.20 & 15.02 & 18.70 & 18.93 & 15.40 & 19.65 \\
& SSIM $\uparrow$ & 0.629 & 0.702 & 0.513 & 0.645 & 0.534 & 0.711 & 0.632 & 0.635 & 0.428 & 0.603 \\
& LPIPS$\downarrow$ & 0.281 & 0.272 & 0.512 & 0.455 & 0.364 & 0.328 & 0.449 & 0.447 & 0.535 & 0.405 \\
[0.8pt] \hdashline \noalign{\vskip 2pt}

\multirow{3}{*}{\makecell[l]{MonoGS~\cite{matsuki2024gaussian}}} 
& PSNR$\uparrow$ & 23.58 & 23.45 & 17.04 & 16.19 & 15.03 & 15.09 & 16.41 & 13.69 & 14.91 & 17.27 \\
& SSIM $\uparrow$ & 0.609 & 0.762 & 0.671 & 0.543 & 0.490 & 0.512 & 0.457 & 0.502 & 0.384 & 0.548 \\
& LPIPS$\downarrow$ & 0.643 & 0.757 & 0.643 & 0.714 & 0.670 & 0.710 & 0.755 & 0.767 & 0.734 & 0.710 \\
[0.8pt] \hdashline \noalign{\vskip 2pt}

\multirow{3}{*}{\makecell[l]{SplaTAM~\cite{keetha2024splatam}}} 
& PSNR$\uparrow$  & 25.51 & 27.40 & 17.84 & 17.10 & 19.24 & 18.30 & 13.68 & 13.17 & 10.01 & 18.03 \\
& SSIM $\uparrow$ & 0.709 & 0.798 & 0.652 & 0.526 & 0.666 & 0.597 & 0.488 & 0.565 & 0.327 & 0.592 \\
& LPIPS$\downarrow$ & 0.214 & 0.235 & 0.346 & 0.448 & 0.295 & 0.359 & 0.456 & 0.380 & 0.517 & 0.361 \\
[0.8pt] \hdashline \noalign{\vskip 2pt}

\multirow{3}{*}{\makecell[l]{Gaussian-LIC}} 
& PSNR$\uparrow$  & \textbf{29.89} & \textbf{31.28} & \textbf{23.90} & \textbf{25.27} & \textbf{22.47} & \textbf{23.49} & \textbf{21.06} & \textbf{22.87} & \textbf{20.73} & \textbf{24.55} \\
& SSIM $\uparrow$ & \textbf{0.820} & \textbf{0.840} & \textbf{0.830} & \textbf{0.805} & \textbf{0.794} & \textbf{0.811} & \textbf{0.693} & \textbf{0.740} & \textbf{0.595} & \textbf{0.770} \\
& LPIPS$\downarrow$ & \textbf{0.155} & \textbf{0.178} & \textbf{0.173} & \textbf{0.212} & \textbf{0.192} & \textbf{0.187} &\textbf{ 0.330} & \textbf{0.266} & \textbf{0.383} & \textbf{0.231} \\

\midrule
\multicolumn{12}{l}{\cellcolor[HTML]{EEEEEE}{\textit{Novel View}}} \\ 

\multirow{3}{*}{\makecell[l]{Gaussian-LIC}} 
& PSNR$\uparrow$  & 29.28 & 30.91 & 23.28 & 24.52 & 22.03 & 22.59 & 20.19 & 22.12 & 19.57 & 23.83 \\
& SSIM $\uparrow$ & 0.799 & 0.830 & 0.811 & 0.785 & 0.776 & 0.777 & 0.639 & 0.693 & 0.531 & 0.738 \\
& LPIPS$\downarrow$ & 0.160 & 0.180 & 0.179 & 0.215 & 0.196 & 0.195 & 0.336 & 0.272 & 0.391 & 0.236 \\
\bottomrule
\end{tabular}
\vspace{-1.8em}
\label{tab:render_evaluation}
\end{table*}

\subsection{Experimental Setup} 
\subsubsection{Implementation Details} 
Our evaluations are conducted on a desktop PC with an NVIDIA RTX 3090 GPU (24 GB VRAM), a 3.2GHz Intel Core i7-8700 CPU, and 32 GB of RAM. For mapping data preparation, we set $N_l$ to 10, $N_k$ to 11, and $N_t$ to 9. For map initialization, we set $o_a$ to 0.1, $o_b$ to 0.7, $N_s$ to $1e5$, and $R$ to $1e4$. For map expansion, we set $\tau$ to 0.99. As for map optimization, we set the loss weighting $\lambda$ to 0.2 and the number $K$ of selected keyframes to 100. All the learning rates for Gaussian attributes are kept the same with the vanilla~\cite{kerbl20233d} but not decayed with schedulers. We share the same learning rate of the affine matrix with~\cite{kerbl2024hierarchical}.
Experiments across all tested scenarios use the same hyperparameters to ensure a fair and comprehensive evaluation.

\subsubsection{Datasets}
We conduct extensive experiments on three public LiDAR-Inertial-Camera datasets, including the FAST-LIVO dataset~\cite{zheng2022fast} and the R3LIVE dataset~\cite{lin2022r} with a solid-state LiDAR, and the MCD dataset~\cite{mcdviral2024} with a mechanical spinning LiDAR. We test 9 sequences among the three datasets, including (1) \textit{hku2} (\texttt{f0}), \textit{LiDAR\_Degenerate} (\texttt{f1}), and \textit{Visual\_Challenge} (\texttt{f2}) from the FAST-LIVO dataset, (2) \textit{hku\_campus\_seq\_00} (\texttt{r0}), \textit{degenerate\_seq\_00} (\texttt{r1}), and \textit{degenerate\_seq\_01} (\texttt{r2}) from the R3LIVE dataset, and (3) segments of \textit{tuhh\_day\_02} (\texttt{m0}), \textit{tuhh\_day\_03} (\texttt{m1}), and \textit{tuhh\_day\_04} (\texttt{m2}) sequences in the large-scale MCD dataset~\cite{mcdviral2024}.
Note that for all tested datasets, we only adopt the left images if the stereo images are provided. The FAST-LIVO dataset and the R3LIVE dataset provide $640\times512$ images, while the MCD dataset has $640\times480$ images.

\subsubsection{Baselines}
Note that when operating in unbounded outdoor environments, depth data from RGB-D cameras with a limited sensing range is often unreliable. Therefore, we first compare our method with the SOTA RGB-only systems, including the NeRF-based approach NeRF-SLAM~\cite{rosinol2023nerf} and the 3DGS-based method MonoGS~\cite{matsuki2024gaussian}. Additionally, since \textit{no open-source radiance-field-based LiDAR-Camera SLAM system is currently available}, we adapt the state-of-the-art RGB-D method SplaTAM~\cite{keetha2024splatam} to operate with LiDAR and monocular images for comparison. Specifically, we merge point clouds from multiple LiDAR scans and obtain dense depth maps through projection, creating pseudo RGB-D images that can be used as input for SplaTAM. For a fair comparison, we adjust its opacity threshold $\tau$ to match ours, allowing more Gaussians in unbounded scenes, which proves to be effective. To evaluate real-time performance, post-processing is disabled for all compared methods.

\begin{figure*}[ht!]
\captionsetup{font={small}}
\caption{Qualitative rendering performance on the sequences \texttt{f2}, \texttt{r0}, \texttt{m0} and \texttt{m2}.}
\centering
    \newcommand{\wratio}{0.17} 
    \newcommand{\imgheight}{2.85cm} 
    
    \makebox[0.01\textwidth][c]{}
    \makebox[\wratio\textwidth]{\normalsize NeRF-SLAM~\cite{rosinol2023nerf}}
    \makebox[\wratio\textwidth]{\normalsize MonoGS~\cite{matsuki2024gaussian}}
    \makebox[\wratio\textwidth]{\normalsize SplaTAM~\cite{keetha2024splatam}}
    \makebox[\wratio\textwidth]{\normalsize Gaussian-LIC}
    \makebox[\wratio\textwidth]{\normalsize GT}
    \\
    \vspace{1mm} 
    \includegraphics[width=\wratio\textwidth,height=\imgheight]{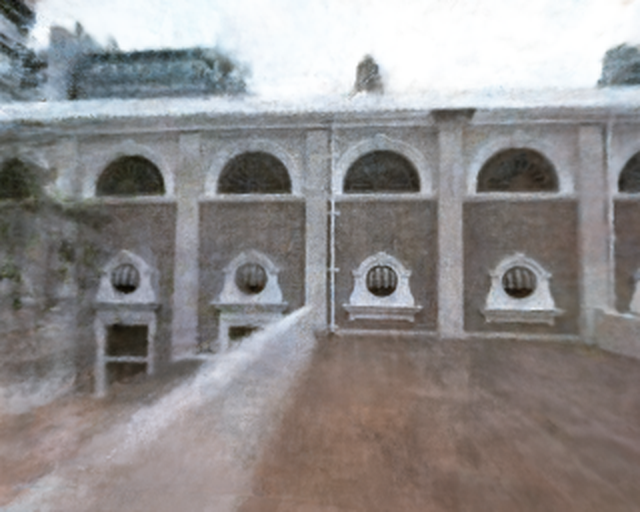}
    \includegraphics[width=\wratio\textwidth,height=\imgheight]{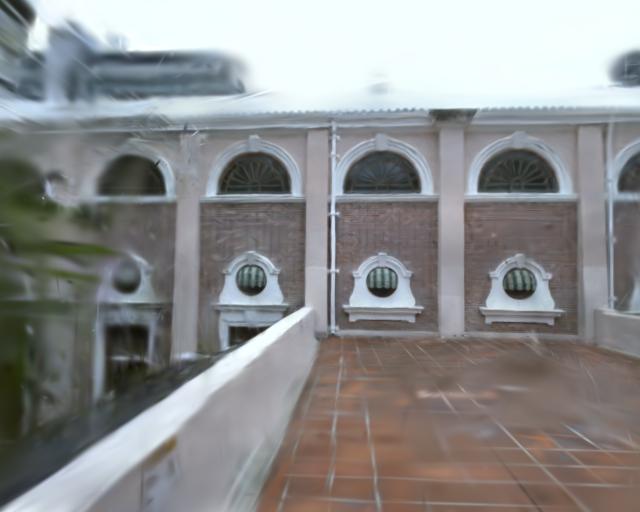}
    \includegraphics[width=\wratio\textwidth,height=\imgheight]{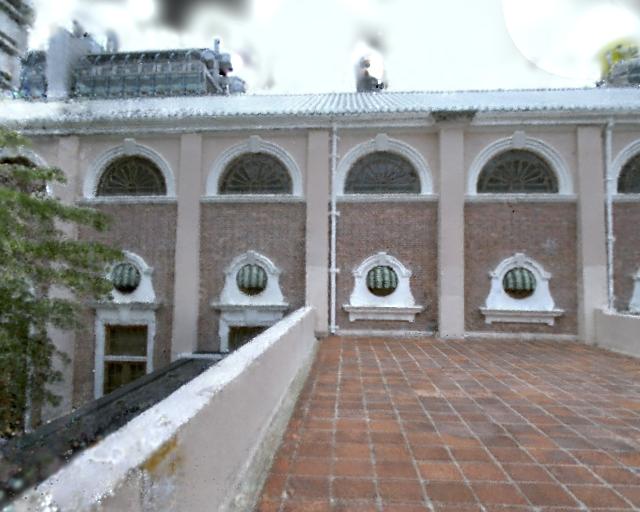}
    \includegraphics[width=\wratio\textwidth,height=\imgheight]{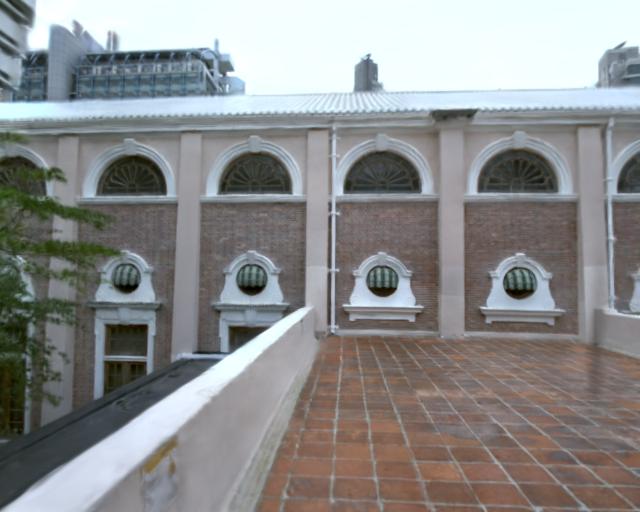}  
    \includegraphics[width=\wratio\textwidth,height=\imgheight]{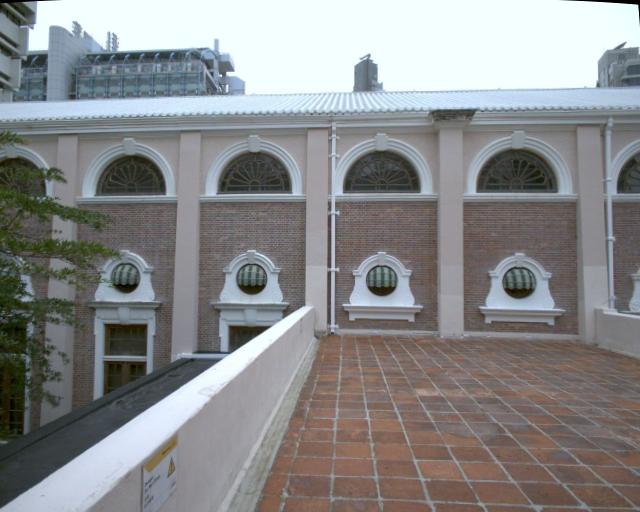} 
    \\
    \includegraphics[width=\wratio\textwidth,height=\imgheight]{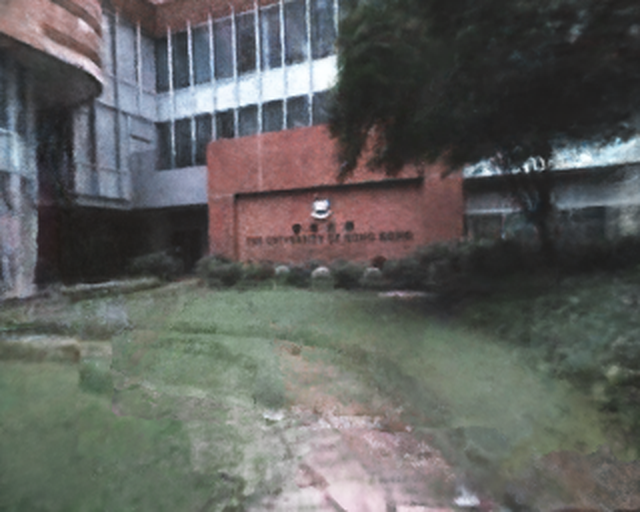}    
    \includegraphics[width=\wratio\textwidth,height=\imgheight]{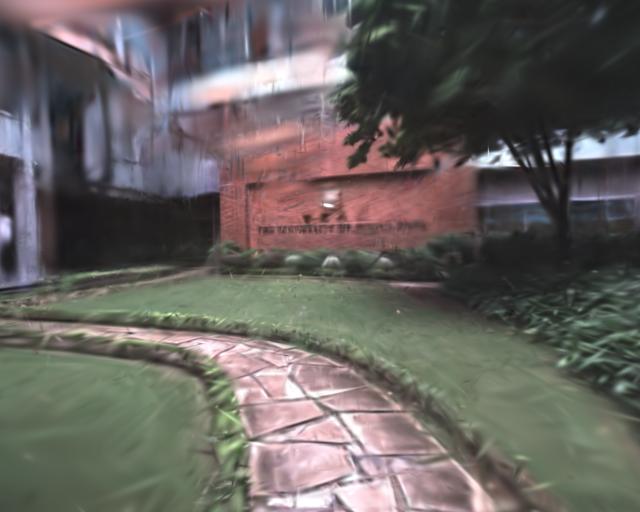}
    \includegraphics[width=\wratio\textwidth,height=\imgheight]{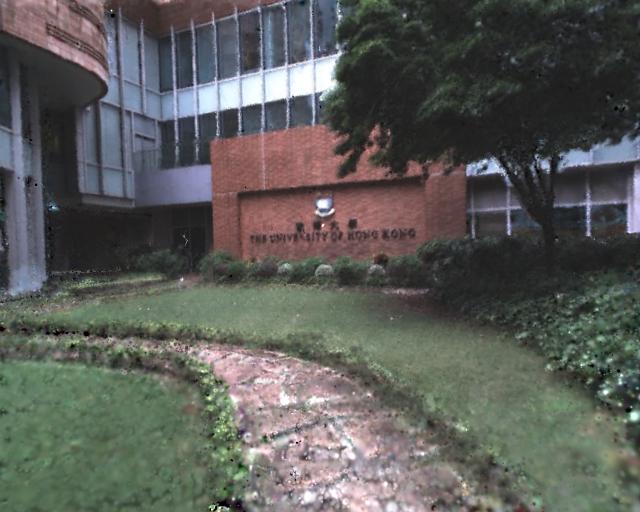}
    \includegraphics[width=\wratio\textwidth,height=\imgheight]{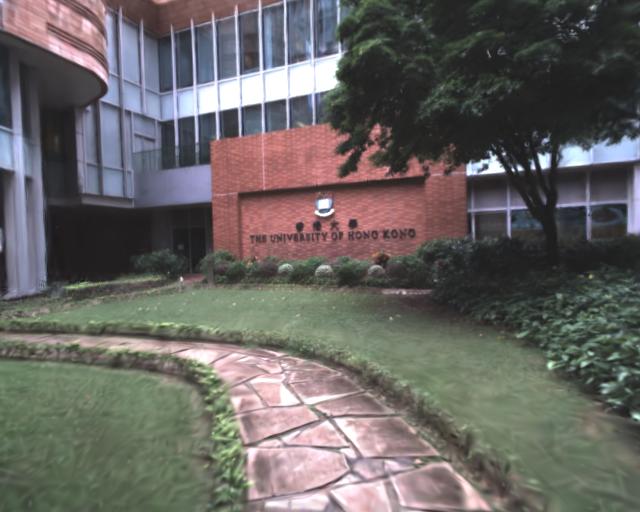}
    \includegraphics[width=\wratio\textwidth,height=\imgheight]{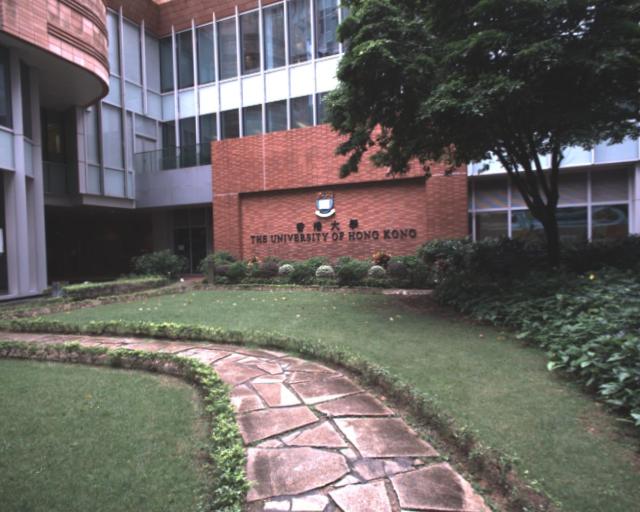} 
    \\
    \includegraphics[width=\wratio\textwidth,height=\imgheight]{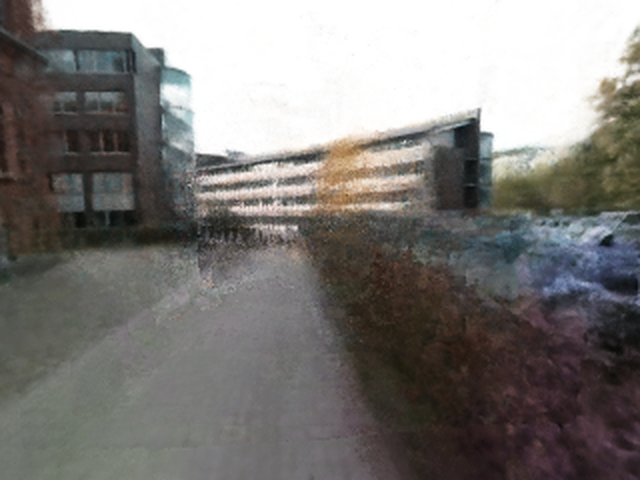}    
    \includegraphics[width=\wratio\textwidth,height=\imgheight]{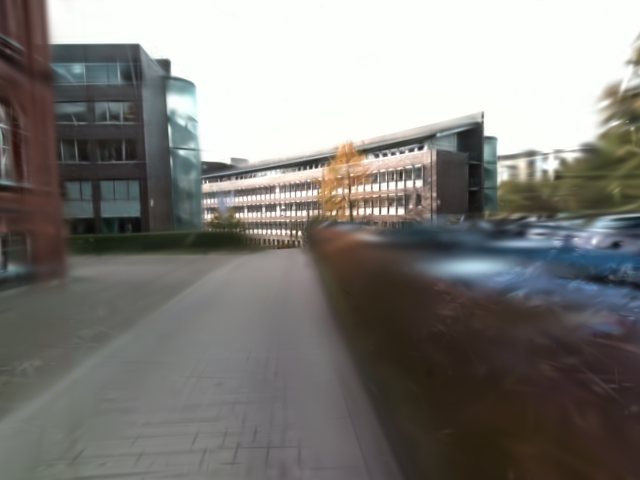}
    \includegraphics[width=\wratio\textwidth,height=\imgheight]{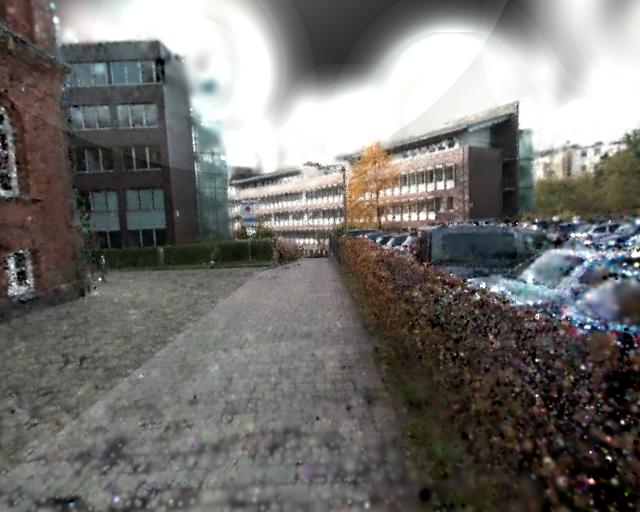}
    \includegraphics[width=\wratio\textwidth,height=\imgheight]{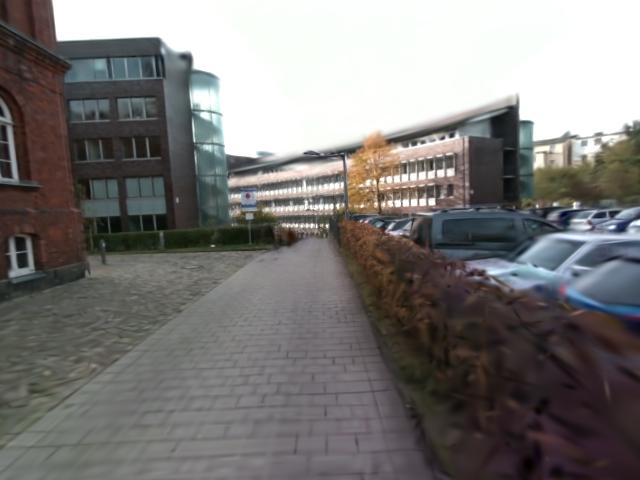}
    \includegraphics[width=\wratio\textwidth,height=\imgheight]{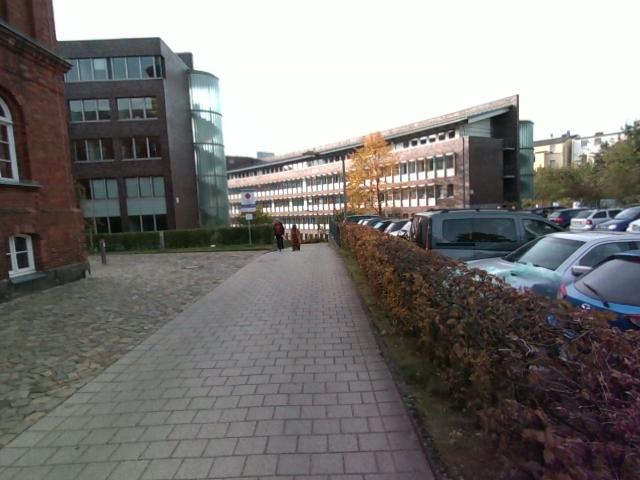} 
    \\
    \includegraphics[width=\wratio\textwidth,height=\imgheight]{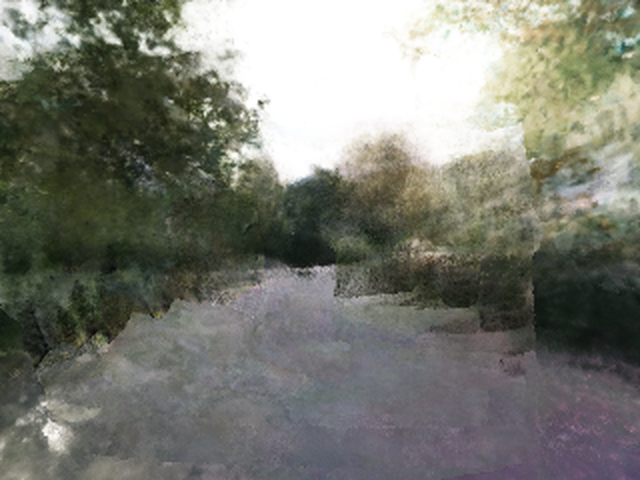}    
    \includegraphics[width=\wratio\textwidth,height=\imgheight]{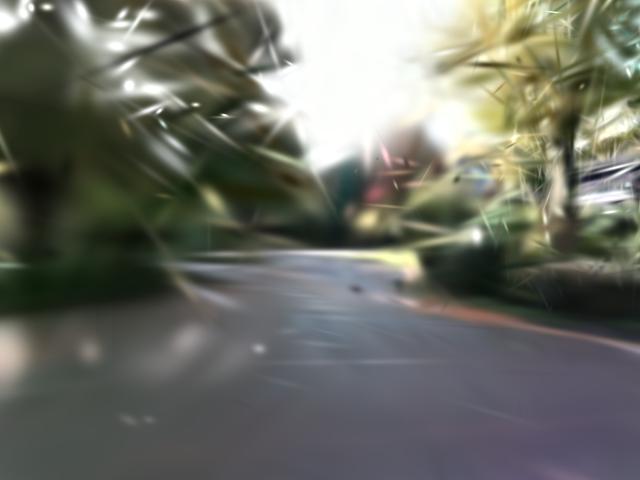}
    \includegraphics[width=\wratio\textwidth,height=\imgheight]{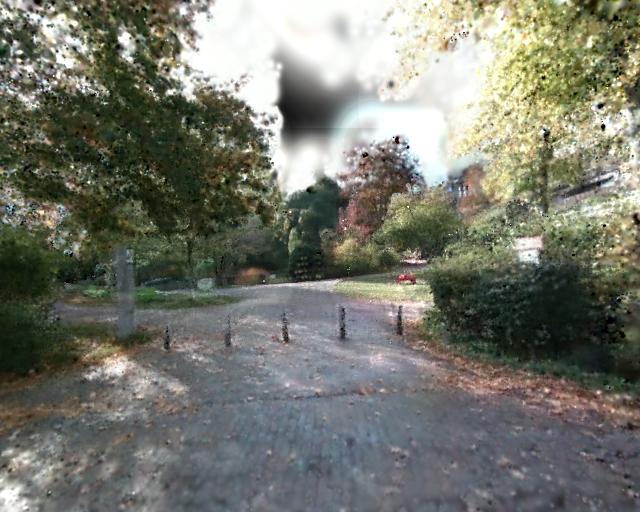}
    \includegraphics[width=\wratio\textwidth,height=\imgheight]{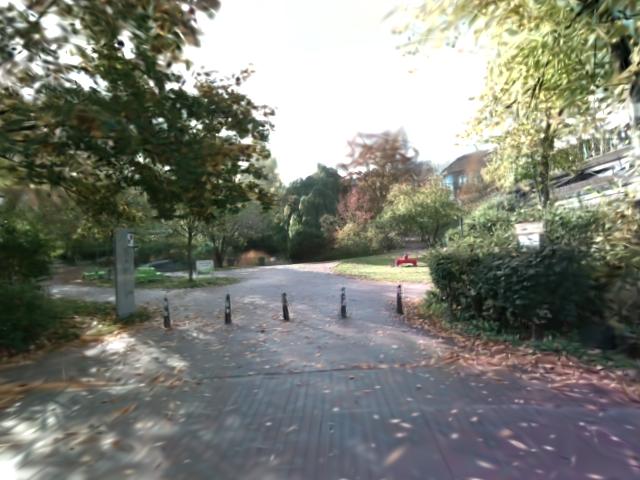} 
    \includegraphics[width=\wratio\textwidth,height=\imgheight]{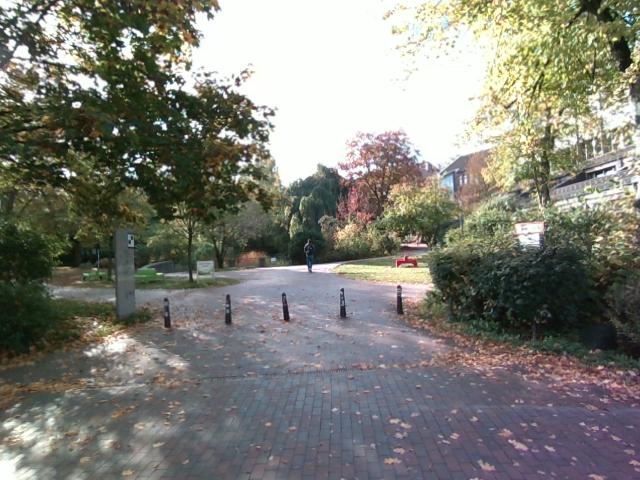} 
    \\
    
    \vspace{-1.4em}
    \label{fig:rendering}
\end{figure*}

\subsection{Evaluation of Mapping}  
\subsubsection{Tracking} 
Based on single-modality sensor input or simple re-rendering loss for tracking, all the compared methods get significant drift or even fail in pose estimation across all the tested sequences. On the contrary, our system consistently achieves superior tracking robustness and accuracy by tightly coupling LiDAR-Inertial-Camera data in a continuous-time factor graph~\cite{lv2023continuous}. 

\subsubsection{Rendering}
We analyze the performance of photo-realistic
mapping by checking the quality of RGB rendering results. Metrics including Peak Signal to Noise Ratio (PSNR), Structural Similarity (SSIM), and Learned Perceptual Image Patch Similarity (LPIPS) are adopted for analysis~\cite{kerbl20233d}. Note that all compared methods struggle with pose estimation in complex scenarios. Therefore, we run them with \textit{ground-truth poses} to purely evaluate their mapping performance. In the FAST-LIVO and R3LIVE datasets, where ground-truth poses are unavailable, we run the compared methods using the more accurate estimated poses from our approach instead. Table~\ref{tab:render_evaluation} shows the evaluation of rendering results on both the training views and novel views. Qualitative results are also shown in Fig.~\ref{fig:rendering}. Lack of precise priors for Gaussian initialization, MonoGS shows degraded performance outdoors. With additional depth supervision estimated from DroidSLAM~\cite{teed2021droid}, NeRF-SLAM achieves slightly improved results but at the cost of significantly increased rendering time. For improved running speed, SplaTAM represents the scene using isotropic Gaussians without considering view-dependent effects, sacrificing visual quality and tending to degenerate in complicated unbounded scenes. With superior geometric priors from a combination of LiDAR points and visual SFM points, and the appropriate modeling of sky and camera exposure, our system consistently achieves the best rendering quality across all sequences on the training views. Even more promisingly, the quality of our renderings at novel views surpasses that of other methods, even on their own training views.

\begin{table}[t]
\centering
\captionsetup{font={small}}
\caption{Runtime (seconds) of different methods on the sequence \texttt{f0} (duration: 105 seconds) with their estimated poses.}
\resizebox{\linewidth}{!}{
\begin{tabular}{@{}ccccc@{}}
\toprule
                                & Tracking$\downarrow$ & Mapping$\downarrow$ & Total$\downarrow$ & \multicolumn{1}{l}{PSNR$\uparrow$} \\ \midrule
\makecell[l]{NeRF-SLAM~\cite{rosinol2023nerf}}            & 105 & 288 & 288  & 25.23                       \\
\makecell[l]{MonoGS~\cite{matsuki2024gaussian}}           & 181 & 387 & 387  & 21.42                       \\
\makecell[l]{SplaTAM~\cite{keetha2024splatam}}       &   954    &   1620    &    2574    &    17.37                        \\
\makecell[l]{COLMAP~\cite{schoenberger2016sfm} + 3DGS~\cite{kerbl20233d}} & - & - & 6764  & \textbf{32.56}                       \\
\makecell[l]{Gaussian-LIC} & \textbf{105} & \textbf{105} & \textbf{105}  & 29.89                      \\
\makecell[l]{Gaussian-LIC w/o acc} & 105 & 198 & 198 & 29.89                       \\ \bottomrule
\end{tabular}
}
\vspace{-2.8em}
\label{tab:runtime}
\end{table}

\subsubsection{Runtime}
We evaluate the real-time performance of all compared methods, defining real-time as completing data processing within the duration of sensor data acquisition, without extensive post-processing. All approaches rely on their estimated poses rather than privileged ground-truth poses. For the sequential system SplaTAM, the total runtime is the sum of the time taken by both tracking and mapping modules, whereas for the parallel systems (NeRF-SLAM, MonoGS, Gaussian-LIC), the runtime is determined by the module with the longer time consumption. Tab.~\ref{tab:runtime} presents the results on sequence \texttt{f0} (105s duration). Powered by real-time LIC odometry and carefully designed strategies to accelerate Gaussian map optimization, our method is the only real-time-capable approach while also achieving the highest accuracy. It completes both tracking and mapping immediately after receiving all sensory data. We highlight the importance of the introduced acceleration strategies (see Sec.~\ref{sec:acceleration}) in Tab.~\ref{tab:runtime} with the entry \textit{Gaussian-LIC w/o acc}. Additionally, we show the results for the vanilla 3DGS~\cite{kerbl20233d} initialized by COLMAP~\cite{schoenberger2016sfm}, where the total runtime includes the offline SFM and Gaussian training. Although the offline, batch-based approach with globally consistent poses and extended optimization outperforms online incremental methods, it is unsuitable for real-time robotic applications.

\subsection{Ablation Study}
\begin{figure}[t]
    \centering
    \begin{subfigure}[b]{0.32\columnwidth}
        \includegraphics[width=\textwidth]{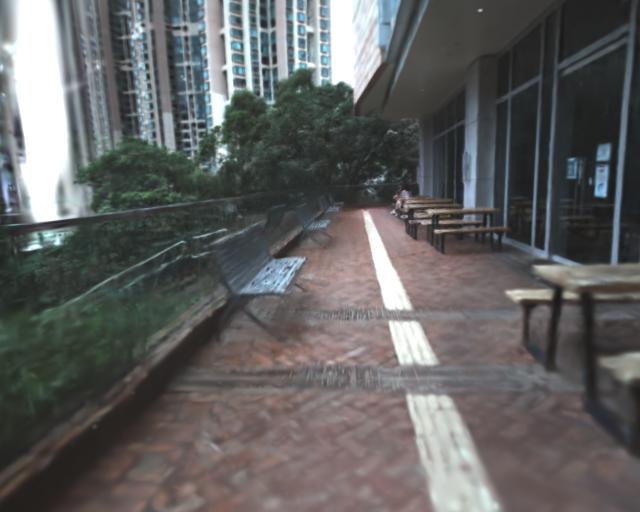}
        \caption{w/o sky modeling}
        \label{fig:sub1}
    \end{subfigure}
    \hfill 
    \begin{subfigure}[b]{0.32\columnwidth}
        \includegraphics[width=\textwidth]{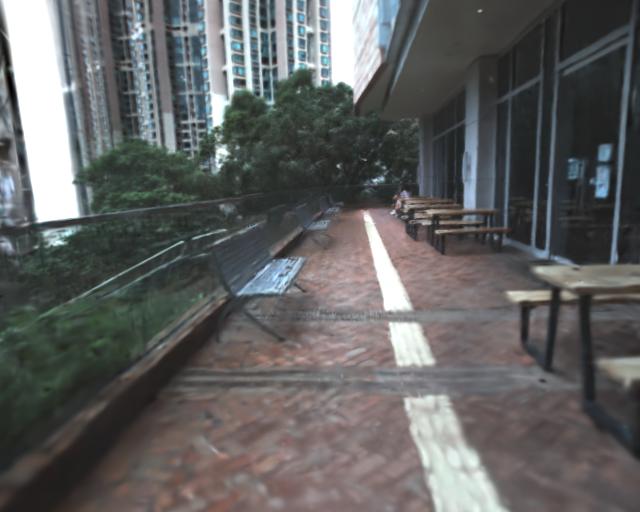} 
        \caption{w/ sky modeling}
        \label{fig:sub2}
    \end{subfigure}
    \hfill 
    \begin{subfigure}[b]{0.32\columnwidth}
        \includegraphics[width=\textwidth]{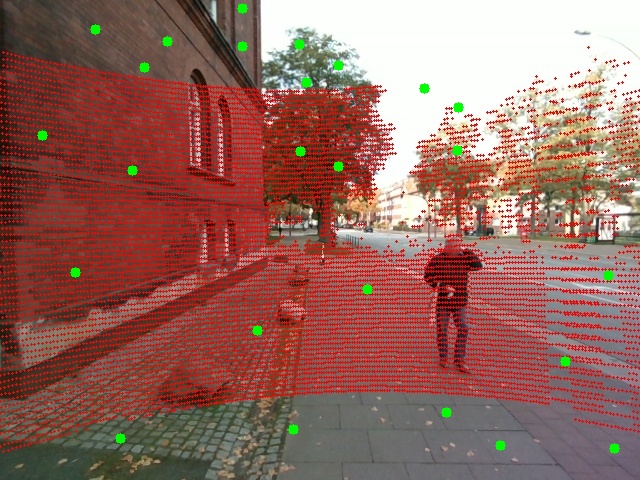} 
        \caption{SFM points}
        \label{fig:sub3}
    \end{subfigure}
    \caption{The significance of sky modeling and incorporating SFM points. (a) and (b) are rendered images on the sequence \texttt{r0}. Without handling the sky, blurring occurs around the buildings (top left) because the surface Gaussians are trying to incorrectly fit the sky. (c) LiDAR points (colored red) cannot cover the entire scene in the camera FoV. We alleviate this problem by additionally incorporating visual SFM points (colored green) obtained by online triangulation.}
    \vspace{-1.7em}
    \label{fig:ablation}
\end{figure}

Table~\ref{tab:ablation} presents the results of the ablation study on our proposed method. The removal of any individual component—such as modeling camera exposures, modeling the sky, or incorporating visual SFM points—leads to a decline in rendering quality. As illustrated in Fig.~\ref{fig:ablation}, the inclusion of sky Gaussians prevents surface Gaussians from incorrectly modeling the sky, thereby enhancing rendering fidelity. Additionally, integrating visual SFM points helps initialize 3D Gaussians in regions where LiDAR data is unavailable.

\begin{table}[t]
\centering
\captionsetup{font={small}}
\caption{Ablation study on the sequence \texttt{f0}.}
\resizebox{0.7\linewidth}{!}{
\begin{tabular}{@{}ccccl@{}}
\toprule
     & w/o expo. & w/o sky & w/o SFM & full  \\ \midrule
PSNR$\uparrow$ & 29.77   & 29.76   & 29.70   & \textbf{29.89} \\ \bottomrule
\end{tabular}
}
\vspace{-2.5em}
\label{tab:ablation}
\end{table}

\section{CONCLUSIONS}
In this study, we introduce a real-time, photo-realistic SLAM system that integrates LiDAR-Inertial-Camera odometry and Gaussian Splatting. Unlike previous radiance-field-based SLAM systems, which are primarily designed for indoor scenes with RGB-D or RGB sensors and struggle in unbounded outdoor scenes or under challenging conditions such as rapid motion and fluctuating lighting, our method is specifically developed to overcome these challenges outdoor. It enhances pose estimation and mapping quality by additionally fusing LiDAR and IMU data. Our SLAM system accommodates both solid-state and mechanical spinning LiDAR and utilizes visual SFM points to initialize Gaussians in areas not covered by the LiDAR. We've also incorporated sky modeling and camera exposure modeling for enhanced rendering. Our system also benefits from optimized strategies for fast Gaussian map optimization. Experiments show its superior performance. In the future, we will try to enhance the accuracy of the odometry by the reconstructed Gaussian map and improve the geometrical reconstruction quality.

\section{ACKNOWLEDGEMENT}
This work is supported by the "Leading Goose" Key R\&D Program of Zhejiang Province in China (2025C01069).

{
\AtNextBibliography{\scriptsize}
\printbibliography
}

\end{document}